\documentclass{nle}

\makeatletter
\let\O@argtabularcr\@argtabularcr
\def\O@xtabularcr{\@ifnextchar[\O@argtabularcr{\ifnum 0=`{\fi}\cr}}
\let\O@tabacol\@tabacol
\let\O@tabclassiv\@tabclassiv
\let\O@tabclassz\@tabclassz
\let\O@tabarray\@tabarray
\def\author@tabular{\authorsize\def\@halignto{}\@authortable}
\let\endauthor@tabular=\endtabular
\def\author@tabcrone{{\ifnum0=`}\fi\O@xtabularcr\affilsize\itshape
 \let\\=\author@tabcrtwo\ignorespaces}
\def\author@tabcrtwo{{\ifnum0=`}\fi\O@xtabularcr[-3\p@]\affilsize\itshape
 \let\\=\author@tabcrtwo\ignorespaces}
\def\@authortable{\leavevmode \hbox \bgroup $\let\@acol\O@tabacol
 \let\@classz\O@tabclassz \let\@classiv\O@tabclassiv
 \let\\=\author@tabcrone \ignorespaces \O@tabarray}
\makeatother

\usepackage{graphicx}
\usepackage[round,colon,authoryear]{natbib}
\usepackage{amsmath,mathtools}
\usepackage{amssymb}
\usepackage{url}
\usepackage{makecell}
\usepackage{textcomp}
\usepackage{xcolor}

\title[]
      {A Text Reassembling Approach to Natural Language Generation}
\author[Xiao Li, Kees van Deemter and Chenghua Lin]
       {Xiao Li, Kees van Deemter and Chenghua Lin}

\received{date 1; Revised date 2}

\pagerange{\pageref{firstpage}--\pageref{lastpage}}
\pubyear{1998}

\begin{document}

\label{firstpage}
\maketitle
\begin{abstract}
Recent years have seen a number of proposals for performing Natural Language Generation (NLG) based in large part on statistical techniques. Despite having many attractive features, we argue that these existing approaches nonetheless have some important drawbacks, sometimes because the approach in question is not fully statistical (i.e., relies on a certain amount of handcrafting), sometimes because the approach in question lacks transparency. Focussing on some of the key NLG tasks (namely Content Selection, Lexical Choice, and Linguistic Realisation), we propose a novel approach, called the Text Reassembling approach to NLG (TRG), which approaches the ideal of a purely statistical approach very closely, and which is at the same time highly transparent. We evaluate the TRG approach and discuss how TRG may be extended to deal with other NLG tasks, such as Document Structuring, and Aggregation. We discuss the strengths and limitations of TRG, concluding that the method may hold particular promise for domain experts who want to build an NLG system despite having little expertise in linguistics and NLG.
\end{abstract}

\section{Introduction}

Natural Language Generation (NLG) is the process of generating natural language texts from non-textual information.
Until approximately the year 2000, NLG research focussed on generating texts from (broadly) logic-based meaning representations, but recent work has generated text from ``flat'' data, such as databases containing huge tables of sensor data, for instance in the weather domain. It is on this ``data to text'' type of NLG that this article will focus.

Different methods exist for performing (data to text) NLG. Among the earliest and most widely used are rule-based approaches (e.g. \cite{reiter2009usi,hunter2011bt,mille2019portable}), which adopt a pipeline to break down the process of natural language generation into multiple subtasks that operate by using manually constructed rules. Another tradition of research focusses on the construction of syntactically structured templates for sentences and clauses, which are then filled and combined to produce texts, in accordance with a set of hand-coded rules \citep{van2005rea,braun2019simplenlg}.

The behaviour of a rule-based NLG system is predictable, because it is governed by deterministic rules that are, to varying extents, readable by humans. When such a system generates textual descriptions for a set of input data, its behaviour and outputs at each stage are fully tracable, as it is essentially a white box model. However, the shortcomings of rule-based approaches are also obvious and well documented. For example, the hand-crafting of rules costs an enormous amount of human effort.
Furthermore, relying on human-crafted rules means that these approaches are almost always highly domain specific: it is difficult to generalise a rule-based NLG system to another domain, which might require a set of completely different rules. In the worst case, the design of the new system may have to be done almost entirely from scratch.

During the last 10-15 years, and especially in the last few years, a growing proportion of NLG research has concentrated on designing a new generation of {\em statistical} NLG approaches~\citep{belz2008aut,konstas2013glo}. These approaches learn knowledge from a data-text parallel corpus and this knowledge is then applied during the generation process. Although this new generation of NLG approaches are, at least in principle, highly generic, and although they can reduce manual NLG labour, they also have their limitations. First of all, as we shall argue below, the above-mentioned NLG systems do not yet completely abandon manually crafted rules.

For example, the system of \citet{belz2008aut} still relies on grammar rules. 
However, when training the system using both training text and a general grammar (such as English grammar), the grammar is often 
unable to cover every sentence structure in the training corpus. 
Consequently, the process of debugging the system involves a lot grammar rule revision. 
Although \citet{konstas2013glo} did not use any grammar rules, their system still requires handcrafted structured descriptions for both the data and the generated texts. This greatly limits the ability of the system to be generalised: since the architecture of the system is based on a specific data structure, it is very difficult to migrate this system to other kind of data. More details of existing statistical NLG approaches will be discussed in Section 2. 

Recent works on statistical NLG has explored the use of Deep-learning \citep[e.g.][]{wen2015sem,li2019stable,duvsek2020evaluating}. Deep learning-based approaches completely discard handcrafted rules and are capable of learning all the knowledge required for generating text from a single training corpus. However, deep neural networks are essentially black box systems; their overall logic is hidden behind large matrices. As such, these NLG approaches lose transparency. It happens frequently that a deep-learning based NLG system produces unexpected or incorrect output \citep{nie2019simple,duvsek2019semantic}, and the way in which these systems are set up means that it can be difficult to modify them in order to fix generation errors. 

We believe that these issues have a considerable impact on the usefulness of NLG in the real world. An approach that does not have transparency (i.e. a black box approach) is difficult to change in light of requirements from clients. For example, a client may want the NLG program to always (or never) output certain sentences under certain circumstances. This is difficult to achieve by programs which have low transparency. It seems very likely to us that this is one of the reasons why statistical approaches have so far had only limited use in practically deployed NLG systems, especially in fault-critical applications. 

Our aim with the work reported in this paper was to develop an approach to NLG that does not require any handcrafted rules, and that is, at the same time, fully transparent. To achieve this aim, we developed the Text Reassembling Generation model (TRG).
Unlike such previous approaches as  \cite{belz2008aut} and \cite{konstas2012uns}, TRG learns to generate Textual Descriptions (TDs, which can be noun phrases, sentences, or multiple sentences) solely based on a standard concept-to-text corpus by reusing the text fragments in the corpus. 
Since the model does not depend on handcrafted rules, domain experts do not have to understand the rules of the language they are generating. Since the model is transparent, domain experts can trace how sentences are generated, and modify the model by hand. We argue that these features of TRG are theoretically desirable and potentially of great benefit to practitioners.

\section{Related Work} 

While many current NLG systems are billed as statistical, we will argue here that many of these these are not completely abandoning the use of handcrafted rules. We also discuss NLG systems based on deep learning, which do abandon all handcrafted rules, but in a way that sacrifices transparency. 

\citep{belz2008aut} introduced a framework called pCRU, which is an end-to-end statistical NLG approach. 
pCRU regards the entire NLG mission as an inversed semantic parsing process, called the expansion algorithm. \citet{belz2008aut}'s approach generates texts for  the original data input  based on a set of Context Free Grammars (CFGs). 
Probabilistic CFGs (pCFGs) are an extension of CFGs, where each CFG is assigned some probabilities indicating its frequency in the corpus. The pCRU framework then generates text following the probability distribution of CFGs. The probabilities, which are learned from a data-text corpus, drive a decision maker. During the generation process, inputs are expanded to grammar trees of the CFGs, and the decision maker governs the expansion process to estimate at which nodes the expansion should terminate. Strictly speaking, the pCRU framework is a semi-statistical NLG approach as it requires handcrafted CFGs for the base generator.
A number of later approaches adopt a similar rationale to Belz's framework, based on a variety of grammatical formalisms, such as the Openccg framework based on Combinatory Categorial Grammar \citep{white2009perceptron,white2012minimal}, and Tree Adjoining Grammar \citep{gardent2015multiple}.

\cite{liang2009lea} introduced an automatic alignment approach, which aims to align the data representation to the texts within a data-text corpus. 
Although not designed for NLG originally, the approach of \citeauthor{liang2009lea} has impacted many NLG approaches as it can produce data and text alignments which are essential for NLG systems \citep[e.g.][]{konstas2012uns,mei2015talk,gatt2018survey,cao2020generating}. 
For example, \cite{angeli2010sim} proposed an approach which takes data to text alignment as input. The alignment information is used for extracting templates at multiple levels of granularities (i.e. sentence templates, expression templates, and word-level templates), and then texts are generated based on the templates in a hierarchical process. 

While evidently very useful, Liang's work has some important limitations. First, it's approach to alignment assumes that each data instance is expressed {\em at most once} in the text. Second, it assumes that each word in the text describes some data; this is problematic for function words (such as ``the'', ``to''), since these link other words, instead of expressing any data by themselves. Consequently, function words have to be treated as part of other content words. In practice, Liang's approach does not provide a principled solution to decide  which text segment (consisting of content words) the function words belong to, e.g., the text segment preceding or after the function words. Importantly, Liang's alignment method assumes that a word can only express one type of data or a single dimension of data (e.g., wind speed or wind direction): it can cannot handle situations in which words express multiple data dimensions, for instance as when the word ``mild" expresses a combination of warm temperatures and low wind speed. These limitations limit the quality of data-to-text alignment.  

\cite{konstas2012uns,konstas2013glo} proposed a statistical end-to-end NLG approach, which extends the automatic alignment approach of \citet{liang2009lea}. Instead of simply using the alignment information, Konstas and Lapata recast the model as a simple probabilistic Context Free Grammar (pCFGs), which 
describes both the structures of the input data and the structures of texts. Then, the pCFGs are packed into a hypergraph; the weights in the hypergraph are learnt through an EM algorithm. In this way, the authors treat sentence generation as finding the maximally likely derivation tree.
As Kanstas and Lapata builds upon Liang's work, it inherits the above-discussed limitations of that method. Their approach is also limited in dealing with data from the domains which are outside the coverage of the pCFG rules. While Kanstas \& Lapata's approach is similar to ours in terms of overall architecture of alignment and generation, our proposed TRG model addresses their limitations without requiring any handcrafted rules.

Recent years have seen a surge of interest in applying Deep Neural Networks (DNN) to NLG related tasks such as dialogue generation~\citep{wen2015sem,li2017adversarial},  referring expression generation~\citep{liu2017referring,yu2017joint}, and language style transfer~\citep{fu2018style,yang2018unsupervised}. 
The general learning paradigm of DNN based approaches for language is to learn some dense, low-dimensional vector representations, which capture the grammatical and semantic generalisations of the input text. 
Such representations, along with some tailored network architectures, are then applied to perform the task of interest. 

Although DNN-based approaches have achieved remarkable success, they have significant drawbacks as well. For instance, deep learning  approaches have been criticised for lacking in transparency \citep{techtalks.tv}, i.e., difficult to interpret and functioning more or less as a black box. 
Consequently, deep learning approaches are difficult to relate to linguistic insights, and they are difficult to modify if and when this is needed.
For instance, it's impossible to manually tweak or update a deep learning based decision support system say, when there is new regulatory changes that need to be taken into account: new training data which incorporates the new regulations would be required for retaining the model. 
Another well-known problem in deep learning approaches is \textit{hallucination}, which is particularly prevalent in the language domain ~\citep{moosavi2017uni,dusek2019eva}. In the context of NLG, hallucination means the generated text contains non-existent or incorrect information of input data. This is of course unacceptable in many real-world applications for instance when generating text that communicate important medical, financial, or engineering information. The recent E2E challenge \citep{dusek2019eva} shows an example whose  input data is ``\texttt{name[Cotto]}, \texttt{eatType[coffee shop]}, \texttt{near[The Bakers]}'', whereas the corresponding text generated by an NLG system was ``\textit{Cotto is a pub near The Bakers}'', confusing a pub with a coffee shop. Another example is \citet{lin2019moel}, which proposed an end-to-end neural model for empathetic response generation. Despite the highly sensitive problem domain, the model sometimes generates potentially harmful responses (e.g., ``\textit{That is great!}'') as a response to someone who talkes about a stressful situation he/she has been through. These problems are avoided by our approach.

\section{Methodology}

\begin{figure}[t] 
\centering\includegraphics[width=3.2in]{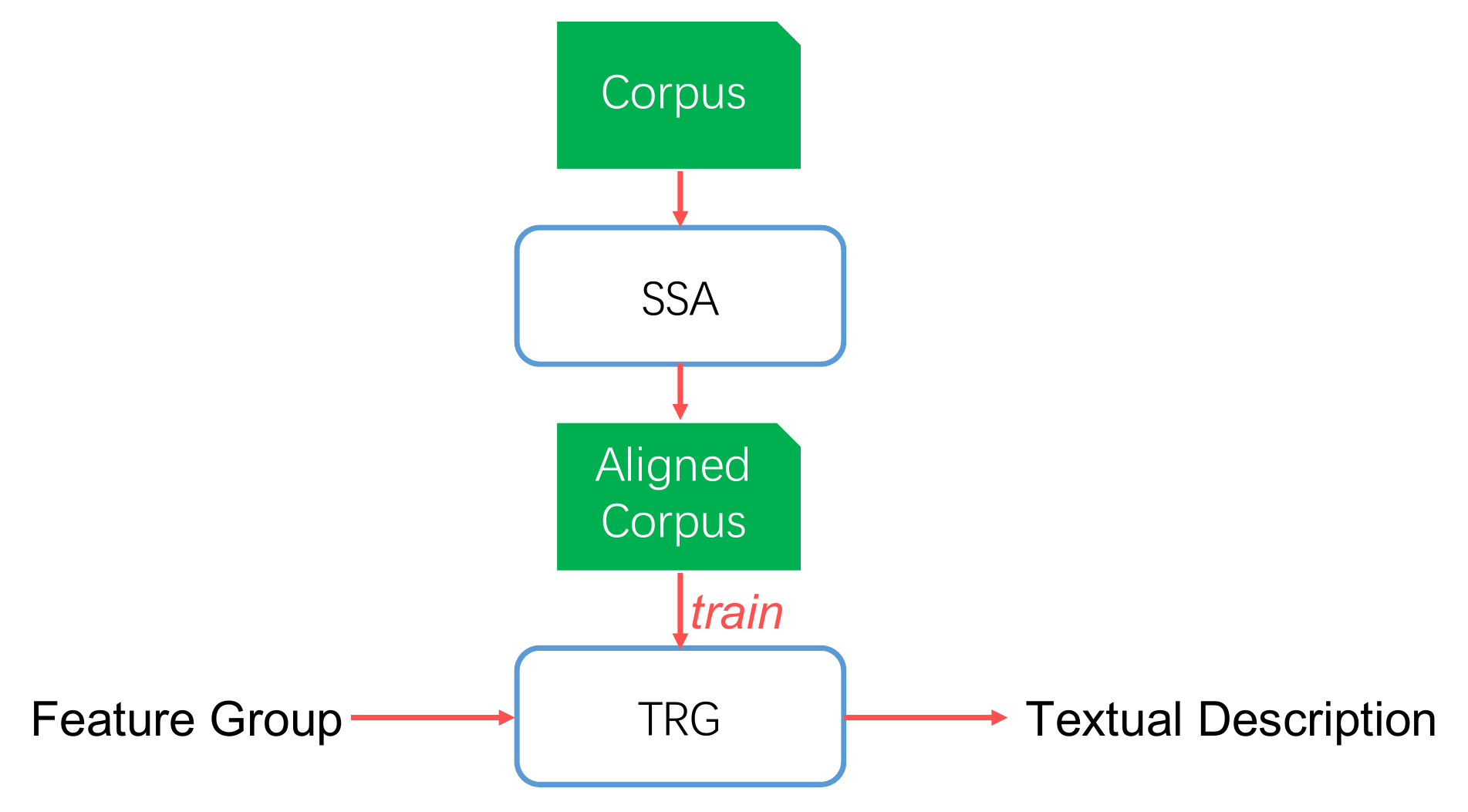} 
\caption{The overall architecture of the TRG model. \textbf{The names in the Figure will be updated soon; SSA should be TRG-Alignment, TRG should be TRG-Generation}}\label{41f1} 
\end{figure}

TRG is divided into two algorithms: %
\begin{enumerate}
    \item \textbf{TRG-Alignment} -- an automatic alignment algorithm which produces a fine-grained alignment between text and data;
    \item \textbf{TRG-Generation} -- a 
    fully statistical NLG algorithm.
\end{enumerate}
Figure \ref{41f1} shows the overall architecture of the TRG model. In a nutshell: {\bf TRG-Generation} accepts data as input, and produces a text as output. Before generation, however, {\bf TRG-Alignment} first has to ensure that the corpus contains a fine-grained alignment between text and data. 
Before explaining the working of these two algorithms in detail, section, let us first make explicit some of our background assumptions and define the main terms that will play a role in the remainder of this section.



\subsection{TRG-Alignment}\label{ssa}

Let us state some assumptions (each of which was made by many authors before us), and define some key notions. 

We assume the existence of what we call a {\em raw data-text corpus}. At the heart of this corpus lie, firstly, a sequence of Textual Descriptions and, secondly, a collection of concepts. A {\em Textual Description} (TD) is a coherent piece of text (for instance, a textual prediction of the weather for a given time interval and a given geographical location), made up of one or more sentences (e.g., ``the temperature is mild''). A {\em concept} is the kind of information that a TD attempts to describe. We assume this information to be represented as a combination of a semantic attribute (e.g., temperature, or wind speed) and a value that this attribute can have (e.g., 20.5 degrees, or 5 knots).

Before TD-alignment, we assume each TD to be aligned with the concepts that it describes. Here is an example of a TD together with the \textit{collection of concepts} (CC) with which the TD is aligned (see Figure \ref{42f1} (a) ).

Note that this alignment is still very course-grained. 
The task of the TRG-Alignment algorithm (algorithm 1 above) is to turn this course-grained alignment into a much more fine-grained type of alignment which we call {\em word-level alignment}. By this, we mean a many-many relation that links each concept in the corpus with that part of the TD that expresses it. More precisely, when a concept $c$ is linked with a text fragment $t$, then $t$ is the smallest contiguous part of the TD that describes $c$. 
Such a contiguous part of text we call a \textit{text fragment}. More precisely, a {\em text fragment} is a sequence of words that appears consecutively in a TD. When TRG is done, the TD above is word-aligned with a CC; one example is Figure \ref{42f1} (b):

\begin{figure}[htbp] 
\centering\includegraphics[width=\linewidth]{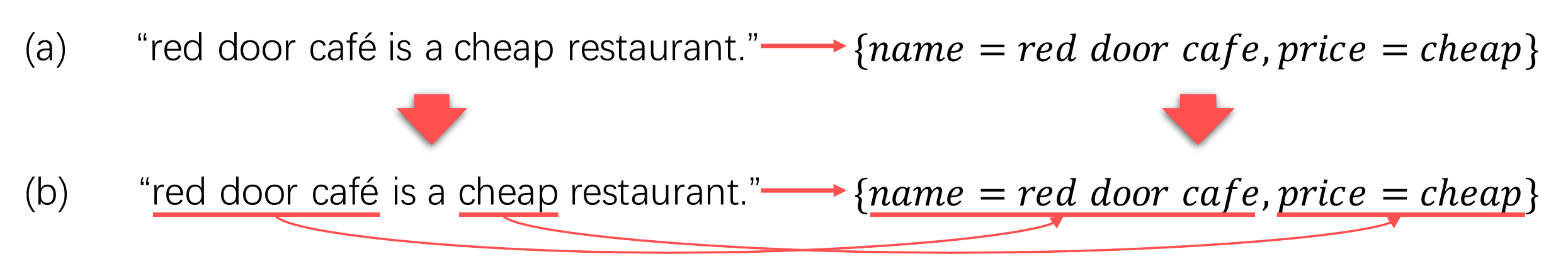} 
\caption{The mapping between TD and CC of a raw concept-to-text corpus}\label{42f1}  
\end{figure}
In Figure \ref{42f1} (b), the TD is split into four fragments, namly ``red door caf\'e'', ``is a'', ``cheap'', and ``restaurant''. The fragment ``red door caf\'e'' is aligned to the feature $name=red~door~cafe$, while ``cheap'' is mapped to $price=cheap$. Other fragments are not aligned to any feature (i.e., they are aligned to the empty set).

TRG-Alignment, as we have seen, relates fragments to concept. As we will explain below, this alignment is a many-to-many relationship, in which multiple fragments can be aligned with multiple concepts. 

Unlike word-to-concept alignment in \citet{liang2009lea}, we can handle two common situations. Firstly, we can handle words that express multiple concepts, for instance as when the word `muggy' expresses information about both temperature and humidity (i.e.,m two different concepts). Secondly, we can handle situations in which non-contiguous text parts (i.e., two text fragments) express the same concept. For instance, in the TD ``list flights from phoenix to san francisco, and arrive sfo before noon'', both ``san francisco'' and ``sfo'' express the feature $to=san~francisco$).

\subsubsection{Fragment and Relationships between Fragments}\label{frag collect}

Before we are getting into details, we define and clarify some key terms. 

\begin{figure}[htbp] 
\centering\includegraphics[width=\linewidth]{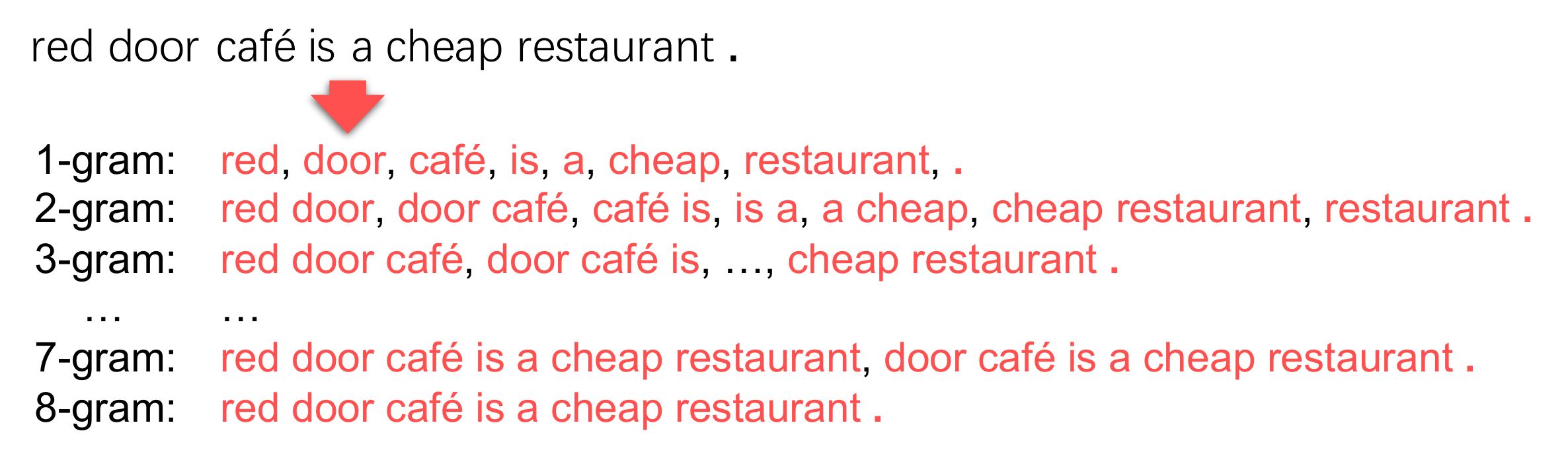} 
\caption{$n$-grams derived from textual description ``red door caf\'e is a cheap restaurant .''}\label{ch4f2} 
\end{figure}

\textbf{Text Fragment} (or {\em fragment} for short). A fragment is a string of contiguous words within a TD, including any punctuation. For example, consider the TD that consist of the sentence ``red door caf\'e is a cheap restaurant.'' This TD consists of 8 ``words"  (including the full stop), so the complete set of all text fragments within this TD has 36 elements (Figure \ref{ch4f2}).

\begin{figure}[htbp] 
\centering\includegraphics[width=\linewidth]{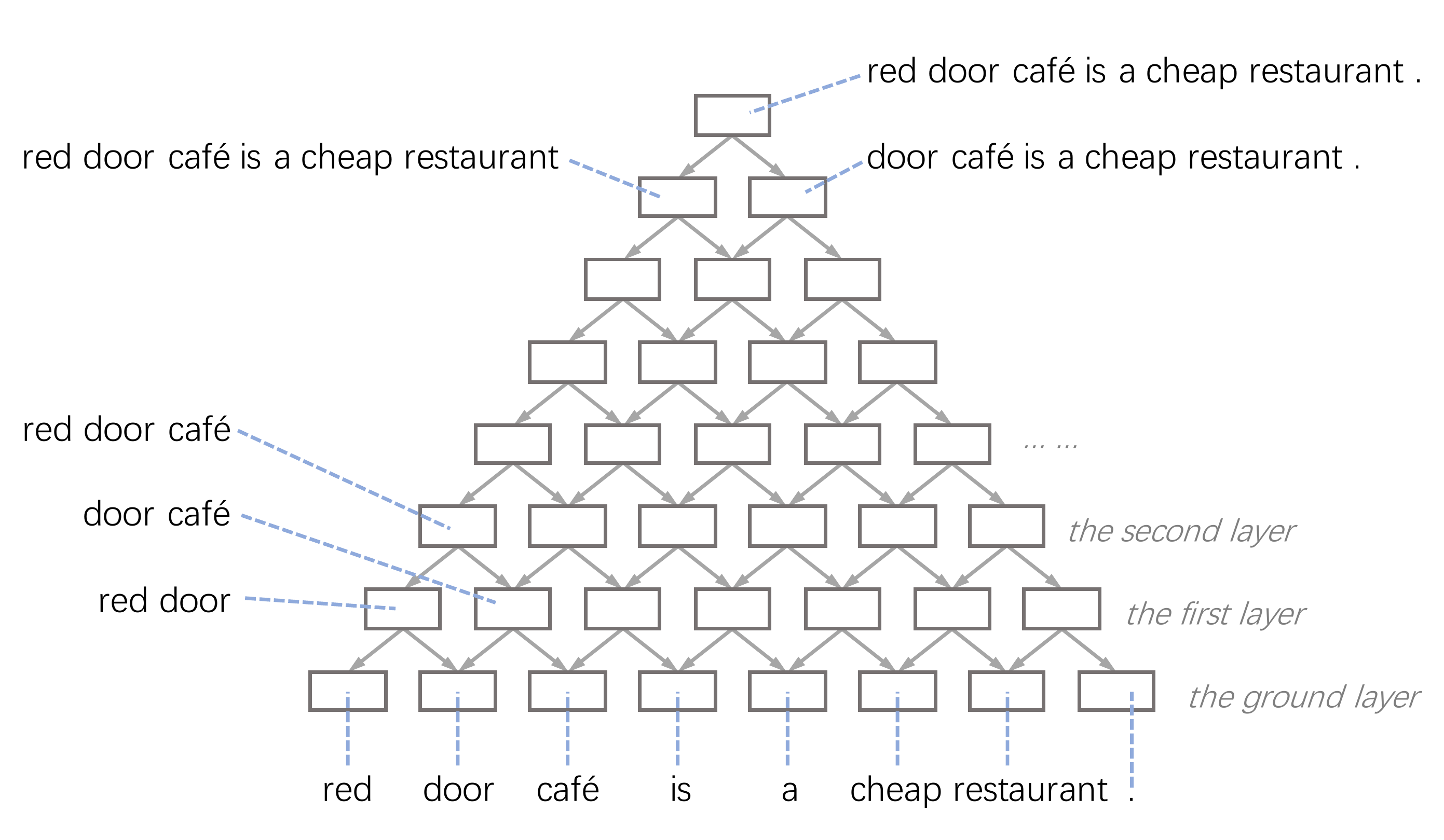}
\caption{The fragment triangle of ``red door caf\'e is a cheap restaurant.''}\label{41f3} 
\end{figure}

 If one fragment ($b$) contains another ($a$) as a part, we say that $b$ includes $a$, denoted by $b\succ a$. Figure \ref{41f3} shows the inclusion relationship between fragments of ``red door caf\'e is a cheap restaurant.''.
This relationship is a lattice, representable as a triangle. The top vertex represents the longest fragment in the TD (i.e. the TD itself) and the bottom vertices are the words (i.e., unigrams). We shall call this the {\em fragment triangle}.

\begin{figure}[htbp] 
\centering\includegraphics[width=5.2in]{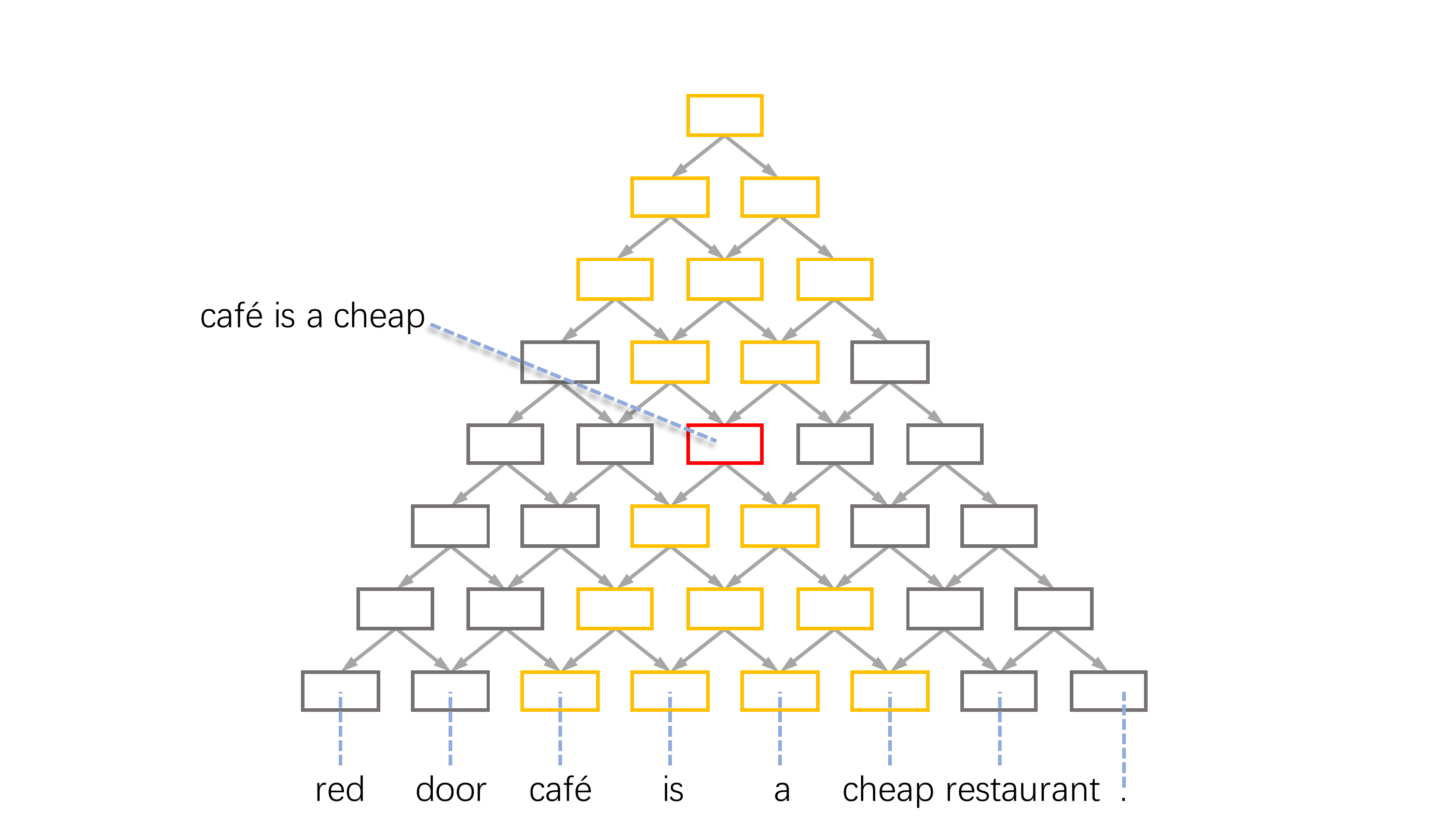}
\caption{The neighbourhood (highlighted by amber border) of fragment ``caf\'e is a cheap''}\label{41f4} 
\end{figure}
We define {\em neighbourhood of a fragment} according to the inclusion relationship. For a fragment $w$, we say fragment $w'$ is the neighbourhood of $w$ if $w'$ includes  $w$  or $w'$ includes $w$. For example, given ``red door caf\'e is a cheap restaurant.'', the neighbours of ``caf\'e is a cheap'' are:
\begin{center}
    ``door caf\'e is a cheap'' \\
    ``caf\'e is a cheap restaurant'' \\
    ``door caf\'e is a cheap restaurant'' \\
    ... \\
    ``is a cheap'' \\
    ``caf\'e is a'' \\
    ``is a'' \\
    ...
\end{center}
The amber boxes in Figure \ref{41f4} are the complete neighbourhood of ``caf\'e is a cheap''.


\subsubsection{Relationships between Fragments and Features}\label{frag-fea}

This section discusses the relation between a fragments ($w$) and features ($g$), which serves as the foundation for our alignment algorithm. We summarise the relationship in two questions:
\begin{enumerate}
  \item Does fragment $w$ expresses feature $g$?
  \item  To what degree does fragment $w$ associate with feature $g$?
\end{enumerate}
These two questions look similar at the first glance but they are different. If a fragment expresses a feature, the fragment represents the semantic meaning of the feature (e.g. ``cheap'' expresses $price=cheap$), but the fragment may also represent other semantic meanings which are not related to the feature. For example, the fragment ``red door caf\'e is cheap'' expresses the feature $price=cheap$, but it also expresses a different feature which is the caf\'e's name. 
We consider both relationships in our alignment algorithm. When comparing two fragments ``red door caf\'e'' and ``red door caf\'e is cheap'',  ``red door caf\'e'' should be highly associated with feature $name=red~door~cafe$, but ``red door caf\'e is cheap'' is less so as it not only expresses the name of the restaurant but also the price. 
This section proposes two functions to present these two kind of relationships.

\noindent\\ \textbf{Modelling Relation, Part 1: Does fragment $w$ express feature $g$?}


The first issue that we face is how to model in terms of probabilities whether a fragment expresses a feature . If a fragment ($w$) expresses a feature ($g$), the conditional probability $P(g|w)$ should be close to 1.
However, using $P(g|w)$ alone is not sufficient to accurately model whether a fragment expresses a feature. That is because if a feature frequently appears in the corpus (i.e.  $P(g)$ is close to 1), for any fragment $w$, $P(g|w)$ will be close to 1.

\begin{figure}[htbp] 
\centering\includegraphics[width=2in]{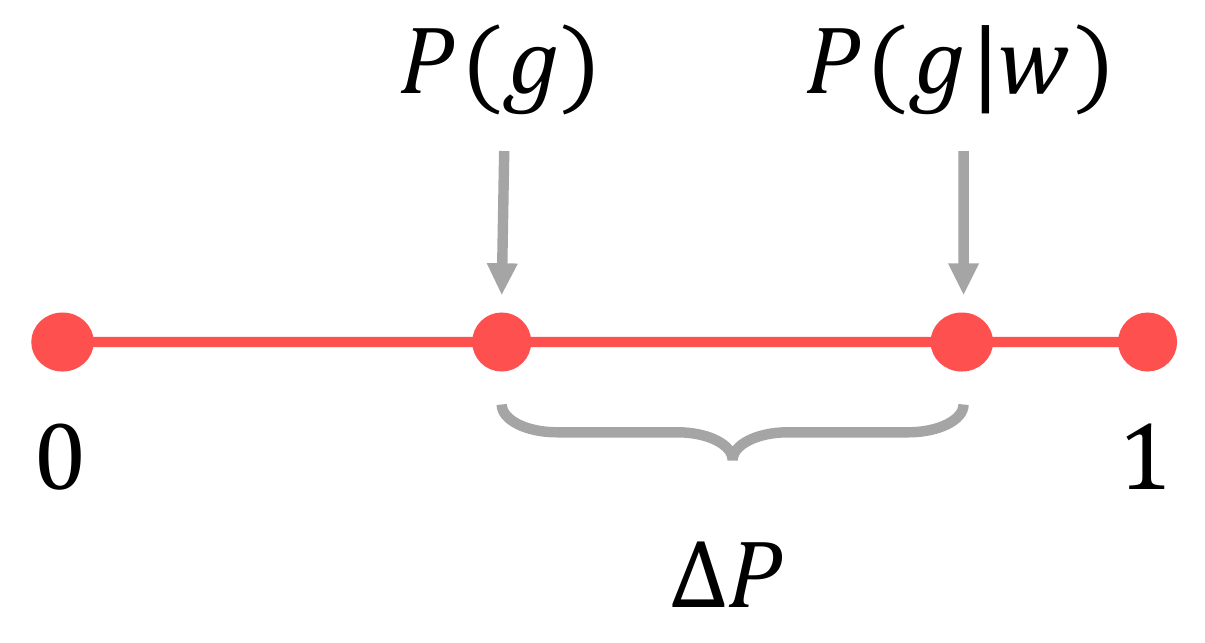} 
\caption{Relationship among values of $P(g)$, $P(g|w)$, and $\Delta P$}\label{42f2} 
\end{figure}

To eliminate this effect, we define a function $Express(w,g)$ to model to what extent a fragment ($w$) expresses a feature ($g$). 
$Express(w,g)$ considers both $P(g)$ and $P(g|w)$; that is, it looks at the differences between $P(g|w)$ and $P(g)$ (i.e., $\Delta P$) as well as how dominant the feature $g$ is in the corpus (i.e., $1-P(g)$):
\begin{equation*}
\begin{split}
Express(w,g) & = \frac{\Delta P}{1-P(g)} \\
 & = \frac{P(g|w)-P(g)}{1-P(g)}
\end{split}
\end{equation*}
We also show an intuitive illustration of the idea behind function $Express(w,g)$ in Figure \ref{42f2}.
%
%

If $P(g|w)$ is smaller than $P(g)$, it means that $w$ does not express $g$.
As probability cannot take negative values, we enforce that $Express(w, g)=0$ when $P(g|w)<P(g)$ (Equation \ref{eq express}).
\begin{equation}\label{eq express}
Express(w,g)=\left\{
\begin{aligned}
& \frac{P(g|w)-P(g)}{1-P(g)} & P(g|w) > P(g)\\
& 0  & P(g|w) \leqslant P(g) \\
& 0  & P(g) = 1
\end{aligned}
\right.
\end{equation}
In addition, if a feature ($g'$) appears in every corpus instance (i.e. $P(g')=1$), the denominator becomes $0$. Therefore we also define that $Express(w, g)=0$ when $P(g)=1$; in other words, if a feature  appears in every instance of a corpus, the feature becomes unimportant or our model estimation.

$Express(w,g)$ can be used to estimate whether a fragment $w$ is a \textit{complete phrase} when $w$ expresses a feature $g$. Here a \textit{complete phrase} refers to a phrase which contains a full specification of feature $g$.  
%
Suppose $a$ and $b$ are two fragments expressing $g$, where $a$ is a complete phrase (e.g., ``red door caf\'e'' or ``red door caf\'e is''), whereas $b$ is an incomplete phrase (e.g. ``caf\'e is'' or ``door caf\'e''), the $Express(a, g)$ must be greater or equal to $Express(b, g)$. When $a$ appears, $b$ must appear because $a \succ b$; but when $b$ appears, $a$ may not appear, because $b$ is an incomplete phrase which can also be part of other phrases used to express other features. For example, ``door caf\'e'' can not only constitute ``red door caf\'e'', but also ``blue door caf\'e'' or ``white door caf\'e''. 
As a result, the probability of $P(feature|incomplete~phrase)$ is lower or equal to $P(feature|complete~phrase)$. In other words, the complete phrase is more likely to express the feature than the incomplete phrase.
For example, when we substitute ``door caf\'e'' and ``red door caf\'e'' into Equation \ref{eq express}, we have
\begin{align*}
    & P(g|\text{``door caf\'e''}) \leqslant  P(g|\text{``red door caf\'e''}) \\
\Rightarrow & Express(g|\text{``door caf\'e''}) \leqslant Express(g|\text{``red door caf\'e''})
\end{align*}


If a fragment ($w$) is not polysemous  ($w$ expresses the same features in any context), $Express(w, g)=1$ where $g$ is any feature that $w$ expresses). That is, $g$ always appears when $w$ appears.
Here we do not consider the case of polysemy, because we assume that a polysemous word can be disambiguated in context. That is to say, if a sentence contains a polysemous word, we only need to find and align fragments which contain both the polysemous word and the necessary context, and this fragment can still be regarded as  non-polysemous.

\noindent\\ \textbf{Modelling Relation, Part 2: To what degree does fragment $w$ associate with feature $g$?}

The function $Express(w,g)$ above estimates the probability of a fragment ($w$) given a feature ($g$), but it cannot identify whether $w$ contains  extra words which express other information. 

To align fragments to features, we identify and align the fragments that only express information about the corresponding feature.  Fragments that express extra information (e.g., information about other features) are not used for alignment. For example, both ``red door caf\'e'' and ``red door caf\'e is'' express $name=red~door~cafe$, but only ``red door caf\'e'' should be aligned to $name=red~door~cafe$.
It should be noticed that we cannot always align the shortest fragment to the feature, because if a fragment expresses a feature, and the fragment contains a rare word, it is possible that the rare word can solely express the feature. Consider the restaurant named ``the stinking rose'' (which expresses $name=the~stinking~rose$). Here ``stinking rose'' contains enough information for $name=the~stinking~rose$, but we should still align ``the stinking rose'' to the feature, because in this case, ``the'' is a part of the name.

Therefore, we define the second function $Core(w,g)$ that adopts the probability $P(w|g)$ to estimate to what degree a fragment $w$ associates with feature $g$ (i.e. in what degree the use of $w$ relies on $g$).
\begin{equation}
Core(w,g)=P(w|g)
\end{equation}
Suppose each of the fragments $w$ and $w_+$ expresses the feature $g$, and $w$ does not contain any extra word which does not express $g$, while $w_+$ does, 
then we have:
\begin{equation}
Core(w,g)>Core(w_+,g)
\end{equation}
Consider the fragments ``cheap'', ``a cheap'', and ``cheap restaurant''; ``cheap'' expresses $price=cheap$, while ``a'' and ``restaurant'' do not. That is, the appearance of ``a'' and ``restaurant'' are events that are independent of the appearance of $price=cheap$. Therefore, we must have:
\begin{align*}
& P(\text{``cheap''}|price=cheap) \geqslant P(\text{``a cheap''}|price=cheap) \\
& P(\text{``cheap''}|price=cheap) \geqslant P(\text{``cheap restaurant''}|price=cheap)
\end{align*}

In addition, if $Core(w,g)=Core(w_+,g)$, the degree to which the extra words in $w_+$ associate with $g$ is equal to the degree with which $w$ associates with $g$. The equality only holds if the extra words always appear when $w$ appears. 
in which case the extra word should be considered as a part of the phrase that expresses $g$. 
For example, in ``the stinking rose'', the article is part of the name. For although ``stinking rose'' by itself can denote the restaurant (i.e. $Express(\text{``stinking rose''},name=the~stinking~rose)=1$), `the stinking rose'' is the fragment that aligns with $name=the~stinking~rose$, because 
\begin{align*}
 &~Core(\text{``stinking rose''},name=the~stinking~rose) \\
 =&~Core(\text{``the stinking rose''},name=the~stinking~rose).
\end{align*}
Whenever ``stinking rose'' appears, ``the'' must also appear.


In another example, ``trattoria contadina'' is a restaurant
(in FS-Restaurant corpus), where both the words ``trattoria'' and ``contadina'' are only used to express the restaurant ($name=trattoria~contadina$) in the corpus. In this case, not only 
\begin{align*}
& ~Express(\text{``trattoria contadina''},name=trattoria~contadina)  \\
= & ~Express(\text{``trattoria''},name=trattoria~contadina) \\
= & ~Express(\text{``contadina''},name=trattoria~contadina)
\end{align*}
but also
\begin{align*}
& ~Core(\text{``trattoria contadina''},name=trattoria~contadina) \\
= & ~Core(\text{``trattoria''},name=trattoria~contadina) \\
= & ~Core(\text{``contadina''},name=trattoria~contadina)
\end{align*}
The values of $Express(,)$ indicate that all the three fragments (``trattoria'', ``contadina'' and ``trattoria contadina'') expresses $name=trattoria~contadina$.
However, the values of $Core(,)$ show all the words associate with $name=trattoria~contadina$, so in the corresponding TD, ``trattoria contadina'' is the true fragment to express the feature, instead of ``trattoria'' or ``contadina''.

Overall, $Core(w, g)$ can test whether a fragment $w$ is the shortest fragment that expresses a feature $g$, when $w$ is known to express $g$. If $w_+$ contains words that do not express $g$, $Core(w|g) \geqslant Core(w_+|g)$; if $Core(w|g)=Core(w_+|g)$, we give up $w$ and regard that $w_+$ truly expresses $g$.

\subsubsection{Aligning Fragments to Features}\label{align frag-fea}






Let us illustrate the alignment process by working through a concrete example. Each record (i.e. TD-CC pair)in the corpus is aligned, calling the functions $Express(w,g)$ and the $Core(w,g)$, each of which takes all data in the corpus into account. 


What fragments align to what features depends on two types of information: whether the fragment expresses the feature, and to what extent the fragment associates with the feature. 
\begin{equation}
\label{eq weight}
    weight(w,g)=Express(w,g) \cdot Core(w,g)
\end{equation}

For each corpus instance (i.e. pair of TD and CC in training corpus), we firstly collect all fragments of the TD, and calculate the value of $weight(w,g)$ for each fragment ($w$) in the collection and each concept ($g$) in the CC. 

%

Consider the FS-Restaurant corpus and focus on the instance in Table \ref{SF sample 1}.
\begin{table}[!h]
\centering
\caption{A real instance in FS-Restaurant corpus}
\label{SF sample 1}
\begin{tabular}{|l|l|}
\hline
\textbf{Textual Description (TD)} & \textbf{Collection of Concept (CC)} \\
\hline
``red door caf\'e is a cheap restaurant .'' & 
$\{name=red~door~cafe,price=cheap\}$ \\
\hline
...&... \\
\hline
...&... \\
\hline
\end{tabular}
\end{table}
The TD expresses the CC with two features $name=red~door~cafe$ and $price=cheap$. So, we assume that fragments of the TD express a (possibly empty) set of features in the feature collection%
We separately substitute $name=red~door~cafe$ and $price=cheap$ into $weight(w,g)$, and for each fragment ($w$) in the TD, we calculate the weights by Equation \ref{eq weight}.

\begin{figure}[thbp] 
\centering\includegraphics[width=5in]{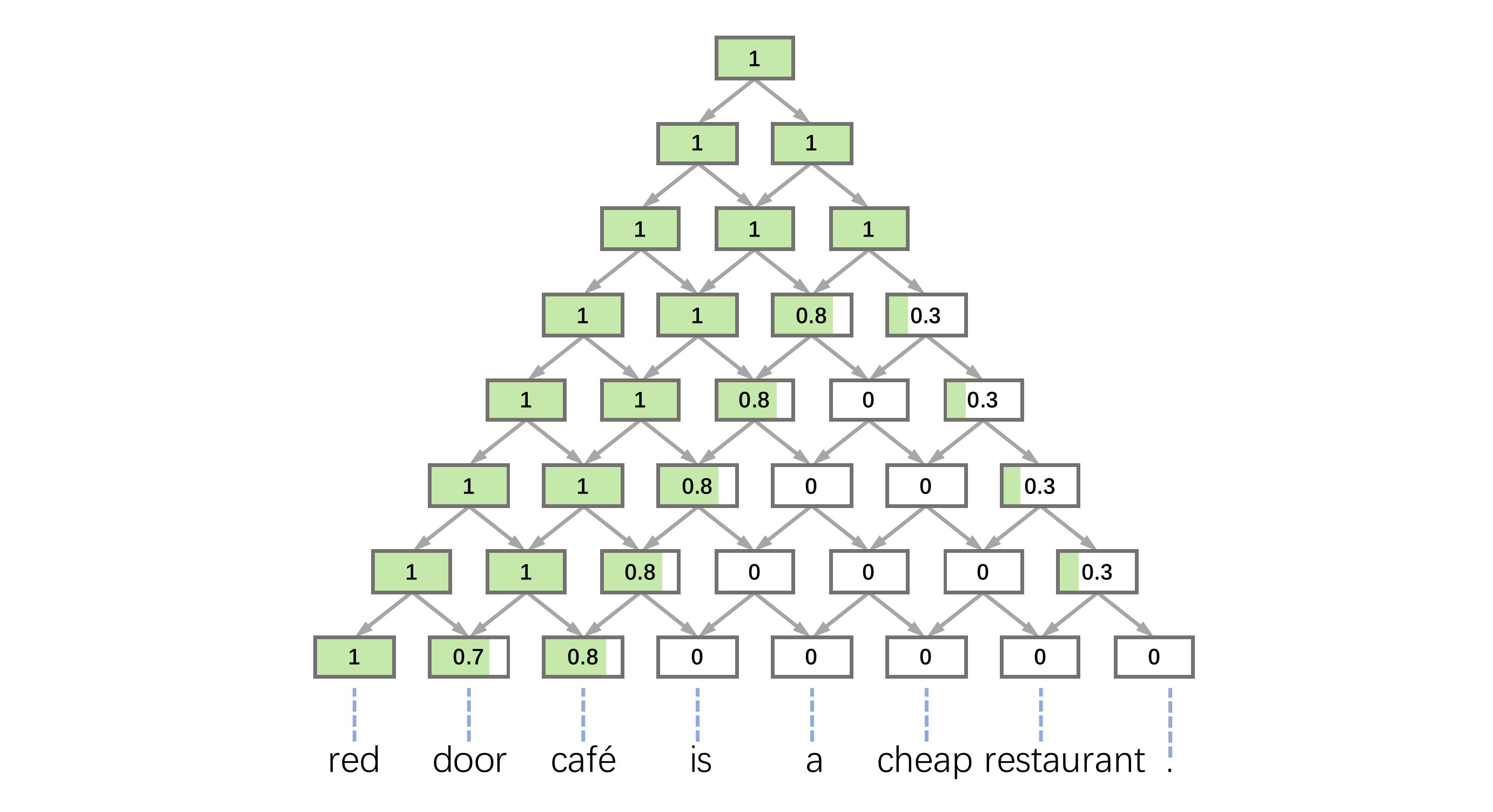} 
\caption{Results of the values of $Express(w,g)$ for each fragment ($w$) in ``red door caf\'e is a cheap restaurant .'' when $g$ is $name=red~door~cafe$}\label{42f4 express} 
\end{figure}

\begin{figure}[thbp] 
\centering\includegraphics[width=5in]{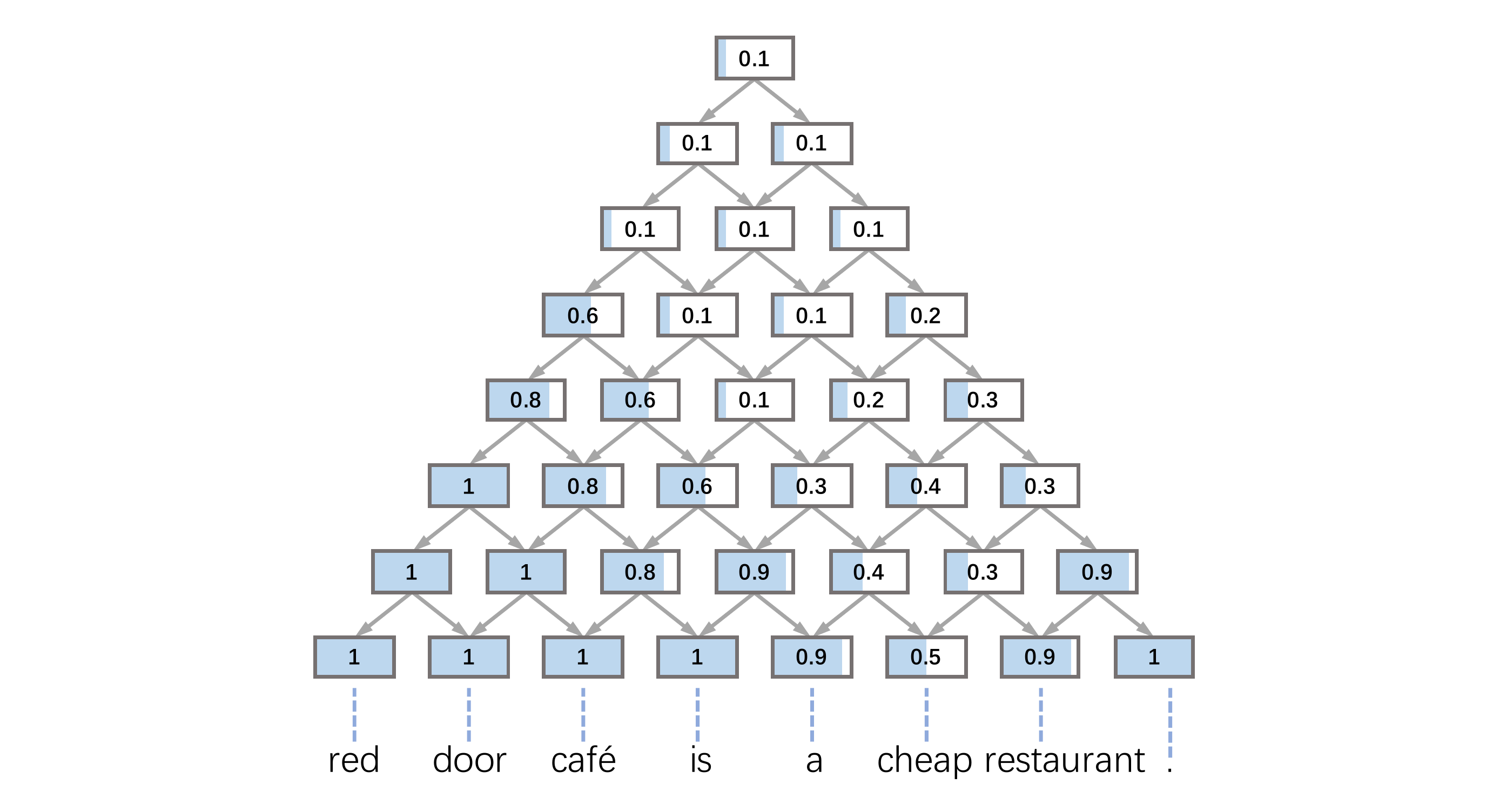} 
\caption{Results of the values of $Core(w,g)$ for each fragment ($w$) in ``red door caf\'e is a cheap restaurant .'' when $g$ is $name=red~door~cafe$}\label{42f4 core} 
\end{figure}

\begin{figure}[thbp] 
\centering\includegraphics[width=5in]{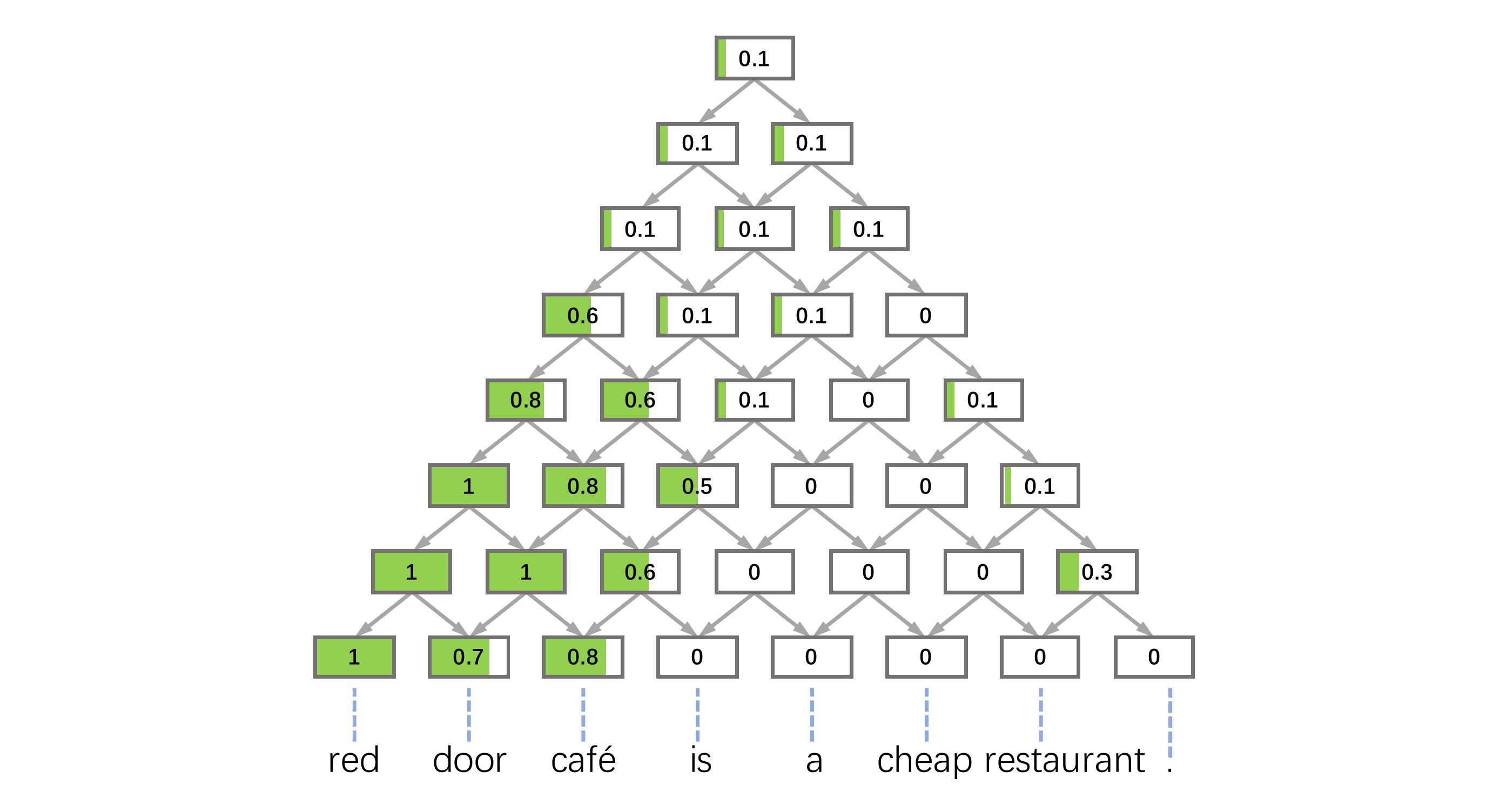} 
\caption{Results of the values of $weight(w,g)$ for each fragment ($w$) in ``red door caf\'e is a cheap restaurant .'' when $g$ is $name=red~door~cafe$}\label{42f4 weight} 
\end{figure}

Figure \ref{42f4 express}, Figure \ref{42f4 core}, and Figure \ref{42f4 weight} show the outcomes of the functions $Express(w,g)$, $Core(w,g)$, and $weight(w,g)$ respectively, given that $g$ presents $name=red~door~cafe$. The function values of each fragment ($w$) is marked in its vertex in each graph. The length of the background colour of a vertex also visualises the function value. 

The outcomes of $Express(w,g)$ (in Figure \ref{42f4 express}) show that any fragment containing ``red door caf\'e'' has a large value (clode to 1), which is in line with our expectations, because any fragment containing ``red door caf\'e'' must express $name=red~door~cafe$. 
Because of the limited size of the corpus, the function values of ``red'', ``red door'', and ``door caf\'e'' are also large. So in this corpus, these three fragments are enough to distinguish (express) $name=red~door~cafe$. But they will not affect our aligning process; the following steps will eliminate this interference.

Comparing to $Express(w,g)$, outcomes of $Core(w,g)$ (in Figure \ref{42f4 core}) show that any fragment which ``red door caf\'e'' contains presents large results, because when the feature $name=red~door~cafe$ appears, these words are always used. At the same time, some other words including ``is'', ``a'', ``restaurant'', and the full stop present large outcomes as well, because they are the frequent words whether the feature appears or not. 

We align the fragments to the feature
according to the outcomes of the function $weight(,)$. If and only if both $Express(w,g)$ and $Core(w,g)$ return large outcomes, $weight(w,g)$ returns large outcome (see Figure \ref{42f4 weight}).
Based on the outcomes of $weight(w,g)$, we focus on the {\em maxima fragments} of the function. Given a feature $g$, a maxima fragment is such a fragment ($w$) that for any neighbour fragment $w'$ of $w$,  $weight(w,g) \geqslant weight(w',g)$.
Since we have marked the values of $weight(w,g)$ on the triangle (Figure \ref{42f4 weight}), the maxima fragments can be intuitively explained through the triangle.
%
\begin{figure}[htbp] 
\centering\includegraphics[width=5in]{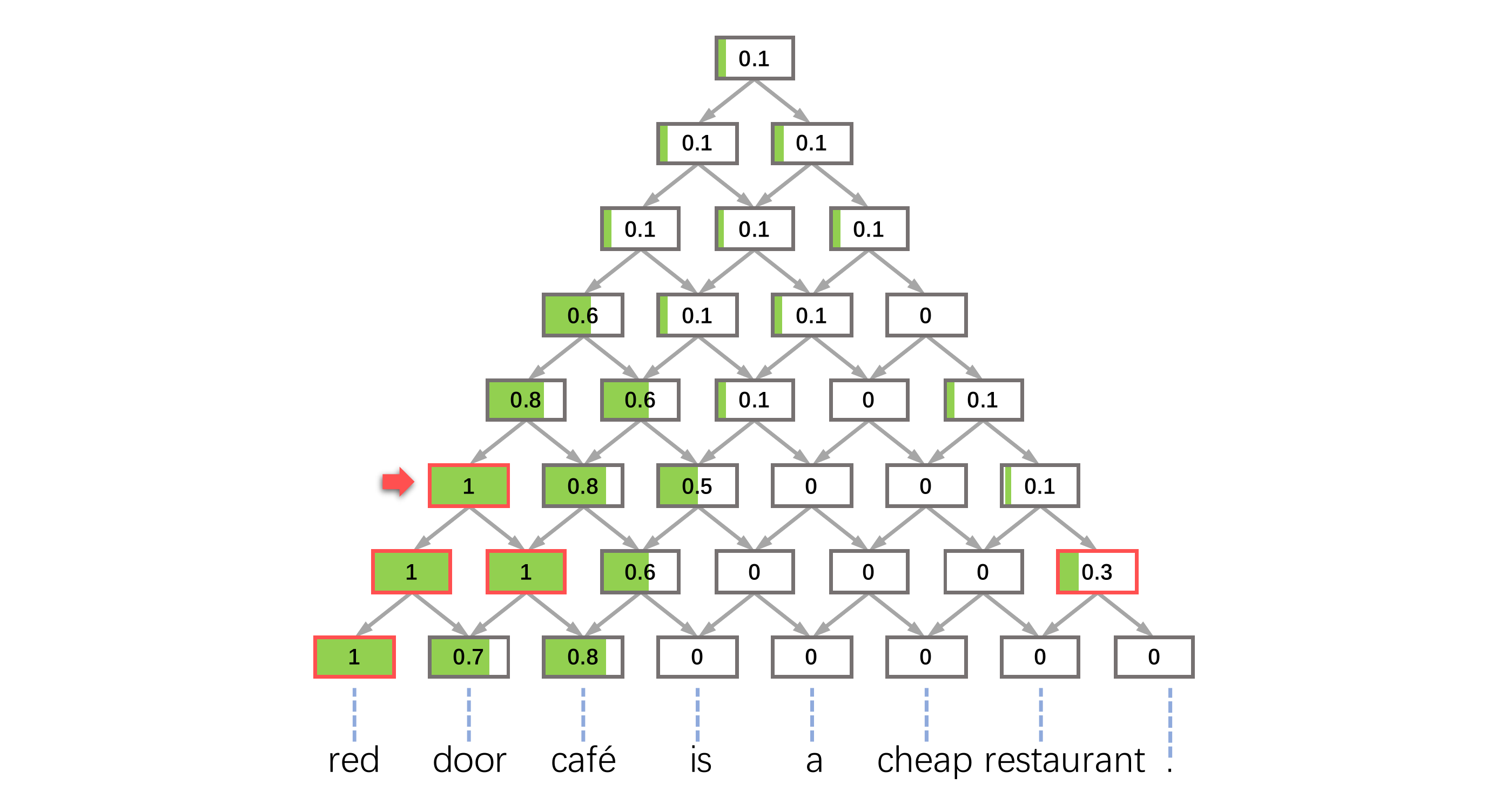} 
\caption{The highlighted maxima fragment region (in red box) for $weight(w,g)$, where $g$ presents $name=red~door~cafe$. The red arrow points the fragment which is aligned to the feature.}\label{42f4 select}
\end{figure}
Figure \ref{42f4 select} highlights the maxima fragments by red border based on Figure \ref{42f4 weight}. 
The maxima fragments may constitute a disconnected sub-graph, but vertices in each connected area of the sub-graph must share the same value (e.g. 1 and 0.3 in Figure \ref{42f4 select}). For each fragment in this area, we select and align the \emph{longest} fragment to the feature.
In addition, we discard the vertices whose value is lower than a pre-decided threshold $\sigma$ (this thesis adopts $\sigma=0.5$) to avoid the effects of noise in the corpus.
Finally, in this case, only the fragment of ``red door caf\'e'' is aligned to $name=red~door~cafe$.

\begin{figure}[htbp] 
\centering\includegraphics[width=\linewidth]{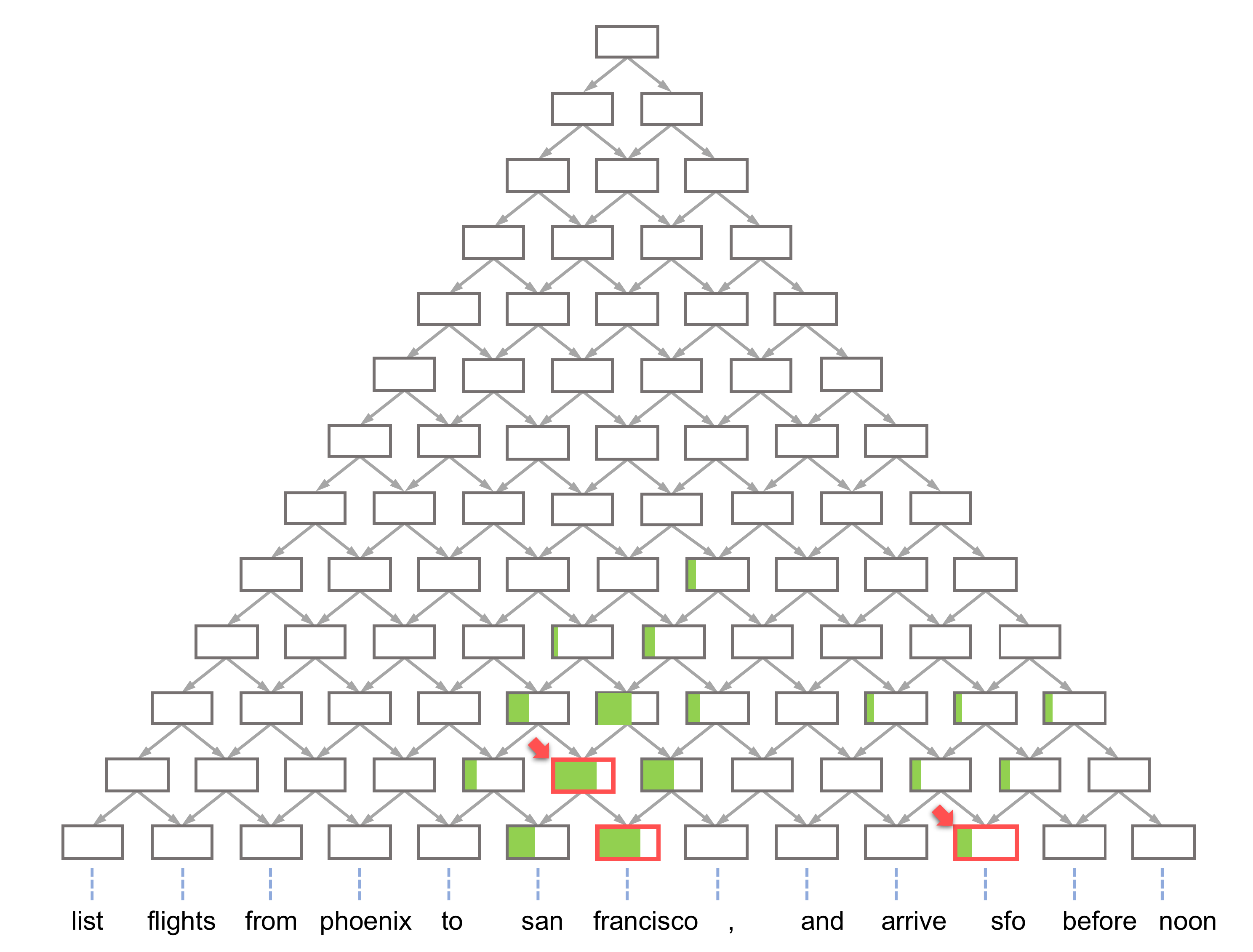} 
\caption{The fragment triangle of ``list flights from phoenix to san francisco, and arrive sfo before noon'' in colour-based marking for the feature $to=san~francisco$. The red arrows point the fragments which are aligned to the feature.}\label{42f8} 
\end{figure}

It is possible that multiple fragments align to one feature, that is, the feature is expressed multiple times in a TD by different phrases. 
Consider the following sentences in the Atis corpus:
\begin{center}
``list flights from phoenix to san francisco, and arrive sfo before noon''
\end{center}
both ``san francisco'' and ``sfo'' express $to=san~francisco$, so, both of them should be aligned to $to=san~francisco$. 
Both of them present maxima values of function $weight(,)$ (see Figure \ref{42f8}).

The fragment triangle shown in Figure \ref{42f8} still uses the same colour-based marking method. The green banners indicate the values of $weight(w, to=san~francisco)$ for each fragment ($w$). The red borders indicate the maxima fragments, and the red arrows point the fragments which are the longest in each disconnected area, and which are selected by the method. It can be seen that this method accurately picked up all the fragments (viz. ``san francisco'' and ``sfo'') that express $to=san~francisco$.

\begin{figure}[h] 
\centering\includegraphics[width=3.2in]{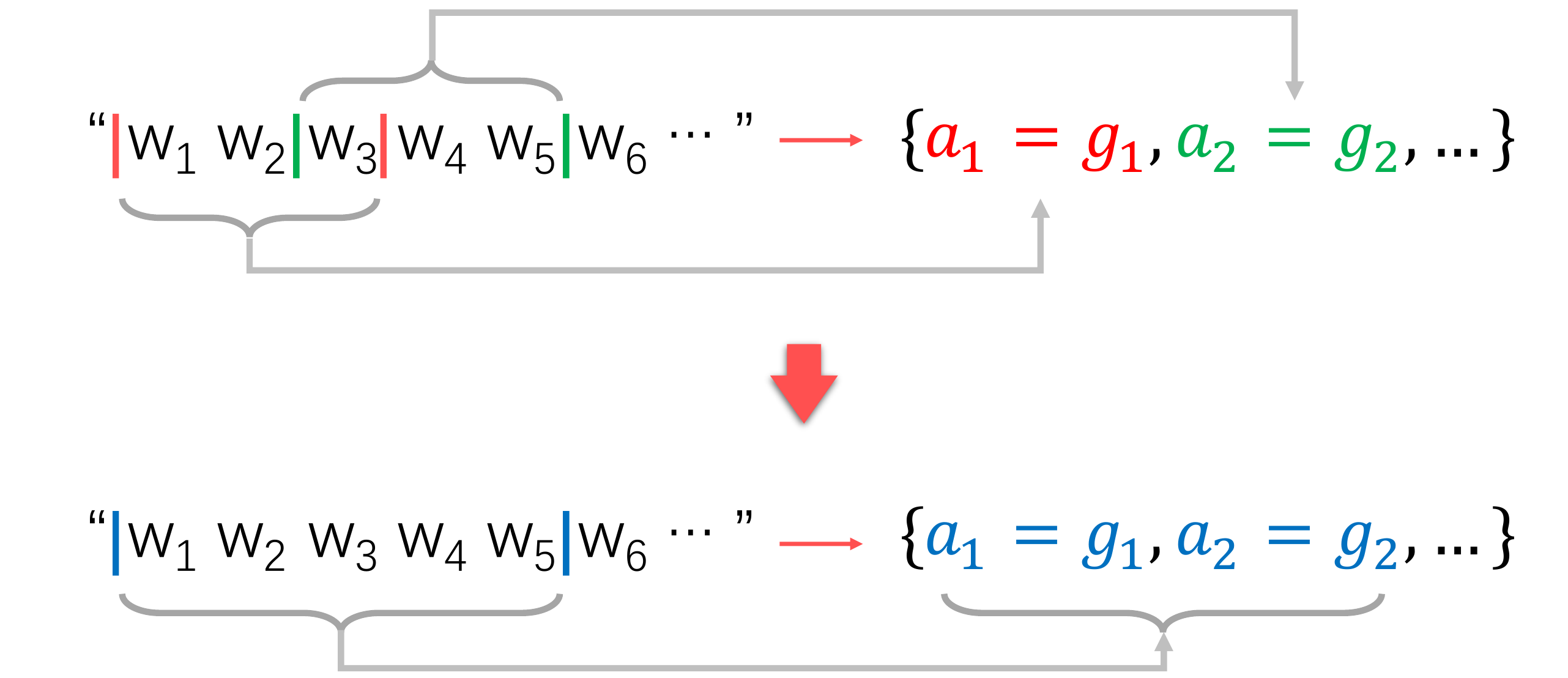} 
\caption{Combining intersected segments}\label{42f9} 
\end{figure}

For each feature corresponding to a TD, we align the fragments to the feature respectively. When the fragments are aligned to all the features, the aligned fragments may be overlapped on each other.
If two fragments overlap, we replace them with their union, and let the union fragment align to both features (see Figure \ref{42f9}).
After the alignment, the TD is split according to the aligned fragments' borders. A segment of the TD can be aligned to one or multiple features (e.g. ``W$_1$ W$_2$ W$_3$ W$_4$ W$_5$'' is aligned to $\{a_1=g_1,a_2=g_2\}$ in Figure \ref{42f9}), and can also be aligned to nothing (e.g. ``W$_6$''). Therefore, the original alignment TD-CC mapping is transformed to the segment/feature mapping.
We apply the method for each corpus instance. After that, the corpus is transformed into the fragment-grained aligned corpus; each TD in the corpus is segmented, and each segment of a TD is aligned to the features that it expresses (except the segments which do not express any feature). 

\subsection{TRG-Generation}\label{trg}

In this section, we introduce the TRG-Generation component. It uses a generation strategy that we call ``splitting-and-reassembling'', which means our generator will operate by reusing fragments produced by the TRG-Alignment component, reassembling them into new sentences.
The TRG-Generation component has it own training process based on the (fine-grained) aligned corpus. It learns (1) what fragments express what features; since TRG-Alignment has already aligned the corpus, learning what fragments express what features is now a relatively simple task. 
(2) It also learns how the fragments constitute a complete TD. To generate linguistically correct sentences, fragments are assembled following a number of \textit{schemata}. We will explain what these schemata are and how they are extracted from sentences (Section~\ref{sc-schema-extraction}), then use the (Section~\ref{sc-selector-train}) extracted schemata to train the generators (Section~\ref{sc-generation}). 
The actual generation process is a two step process: TRG-Generation (1) selects a schema as the sentence plan, then (2) selects fragments to express given features and fills the schema with the fragments.

\subsubsection{Schema Extraction}\label{sc-schema-extraction}

We suppose that sentences organise their information according to some latent structure -- the \textit{schemata}. Each schema restores a possible sequence that sentences can use to express information. Considering the example of the TD below:
\begin{center}
``red door caf\'e is a cheap restaurant.''
\end{center}
This TD expresses three pieces concepts: $name=red~door~cafe$, $price=cheap$ and $type=restaurant$ by the fragments of ``red door caf\'e'', ``cheap'', and ``restaurant'' respectively. We can regard the remaining fragments (viz. ``is a'' and ``.'') express the unknown concepts, because such concepts are not labelled in the corpus. Inspired by this, we convert sentences into the sequence of the concepts that each fragment expresses, and we call this concept sequence the \textit{schema}, for example, the schema of ``red door caf\'e is a cheap restaurant.'' is:
\begin{center}
[$name$] ``is a'' [$price$]  [$type$] ``.''
\end{center}
In schema (like the one above), we use \textit{placeholders}, which are denoted by the square brackets, to indicate the attributes of concepts that the corresponding fragment expresses in each position. For example, [$name$] is a placeholder denoting that the sentence expresses the restaurant name at the beginning position. At the second position, it is a fragment -- ``is a''. Because we do not know what concepts it expresses, we leave the fragment in the schema. 

In this way, a schema can always indicate the order in which a sentence expresses concepts. Although the placeholders only indicate the attribute (e.g. $name$) of concepts, while a full concept is the pair of attribute and value (e.g. $name=red~door~cafe$). However, when a schema is paired with a CC, the missing concept values can be found in the CC.

Converting all the corpus TD into schemata can derive two types of datasets -- \textit{schema dataset} and \textit{fragment dataset}. The schema dataset consists of the pairs of schema and CC. When a corpus TD is converted into a schema, the schema can naturally inherit the CC of this TD, and it is natural that the schema dataset contains such schema-CC pairs.

Fragment dataset consists of fragment-CC pairs.
In particularly, the fragment expresses the paired CC, so that usually a CC in fragment datasets is a sub-collection of a CC paired with a TD. For example, suppose the CC paired with ``red door caf\'e is a cheap restaurant.'' is:
\begin{center}
\{$name=red~door~cafe, price=cheap, type=restaurant$\}
\end{center}
Since the fragment ``red door caf\'e'' expresses $name=red~door~cafe$, there should be a fragment dataset containing the following fragment-CC pair:
\begin{center}
(``red door caf\'e'', \{$name=red~door~cafe$\})
\end{center}
Recall that CC is a collection, so if a fragment expresses multiple concepts, the CC of the fragment-CC pair contains multiple elements.

Since a fragment may express different CCs (e.g. polysemy), we extract a fragment dataset for each placeholder in each extracted schema; that is, a corpus derives a group of fragment datasets, while each fragment dataset only contains the fragment-CC pairs for a specific schema position.

Fragment datasets are obtained when converting corpus TD into schemata. When we convert a TD into a schema, we collect the corresponding relationship of the fragment and the CC that each placeholder derives. When all the TDs are converted into schemata, we aggregate the collected fragment-CC pairs into a fragment dataset if the pairs are derived from the same placeholder. Two placeholders are considered as the same if and only if they are at the same position of the same schema, and two schemata are considered as the same if and only if they are equal at the sequence-level.

Overall, this step, the schema extraction, derives several datasets from converting all the corpus TD into schemata -- one schema dataset and a group of fragment datasets. These two types of datasets will be used to train the TRG-Generation components.

\subsubsection{Training Process}\label{sc-selector-train}

The TRG-Generation component adopts a two-step generation strategy. Firstly, it selects a schema, then, fill the placeholders of the schema by fragments. Both the two steps can be concluded as the same selection task -- given a group of features and a group of candidate items, selecting an appropriate item to express the given features.

TRG-Generation component consists a series of selectors.
The first selector is for schema selection. The schema selector is trained with the schema dataset. Given a feature collection, it selects a schema from the training dataset.
The remaining selectors are the fragment selectors. Each fragment selector is trained by a FV dataset for a specific placeholder.

The schema selector and the fragment selectors share the same architecture and the training process. Here, we employ the selection method of \cite{li2016sta}.
Suppose a training dataset contains $N$ records, we represent each candidate item (the $i$-th item) by a multi-hot vector $\mathbf{s}_i^T$. The $r$-th entry of $\mathbf{s}_i^T$ is 1 if the $r$-th record derives the item $j$, otherwise 0. The dimension of $\mathbf{s}_i^T$ is $N$.
Meanwhile, we represent the feature collection or each record (the $r$-th feature collection) as a multi-hot vector $\mathbf{k}_r$ that each entry of $\mathbf{k}_r$ corresponds to a feature (recall that features are the concept-value pair e.g. $price=cheap$ or $type=restaurant$). The $f$-th entry of $\mathbf{k}_r$ is 1 if the feature collection contains the feature $f$, otherwise 0. The dimension of $\mathbf{k}_r$ is the total number of features (denoted by $M$). Then, all the $\mathbf{k}_r$ constitutes a $N-by-M$ matrix $\mathbf{K}$ whose row vectors correspond the feature collection of all the records.

For each candidate item $i$, we aim to find a mapping vector $\mathbf{p}_i^T$ that given a feature collection (i.e. $\mathbf{k^*}$), $\mathbf{k^*}\cdot \mathbf{p}_i^T$ returns a \textit{selection weight} $w_i$. The value of $w_i$ fits the probability that candidate item $i$ should be selected. 
\begin{equation}\label{Eq_train}
\mathbf{k^*} \cdot \mathbf{p}_i^T = w_i
\end{equation}
Such $\mathbf{p}_i^T$ can be estimated by Least Squares (Equation \ref{Eq_train}, where $pinv(\mathbf{K})$ means the pseudo-inverse of $\mathbf{K}$).
\begin{equation}\label{Eq_train}
\begin{split}
&\mathbf{K} \cdot \mathbf{p}_i^T = \mathbf{s}_i^T \\
\Rightarrow~&\mathbf{p}_i^T = pinv(\mathbf{K}) \cdot  \mathbf{s}_i^T
\end{split}
\end{equation}

We independently train the schema selector and each fragment selector. After training, each selector is equipped an group of mapping vectors learnt from the corresponding training corpus.

\subsubsection{Generation with Probabilities}\label{sc-generation}

Given a feature collection, the most direct generation strategy is to select a schema with the highest probability, then, select fragments (with the highest probabilities as well) for each placeholder. Finally we replace the placeholders by the selected fragment, so the schema is transferred back to a TD.

However, both the schema selection and the fragment selection rely on probabilities. In order to generate the overall optimal TD, we need to consider the weights of the schema selection and the fragment selection of each placeholder as a whole. 

What we are pursuing here is to generate sentences with no linguistic error as much as possible. That is, we want to avoid the situation that when we select a schema (with the highest probability), the schema contains a placeholder that we do not have a appropriate fragment to fill. By this reason, we define the \textit{appropriation weight} $P(x)$ for a generated TD $x$ as the minimum of the selection weights of the selected schema and the selected fragments for each placeholder of the schema.

\begin{equation}
P(x)=Min(w_{schema}, w_{fragment_1}, ..., w_{fragment_n})
\end{equation}
Based on this definition, the most appropriated TD is the one with the highest appropriation weight when given a feature group $\mathbf{k^*}$.
\begin{equation}
x = \underset{x}{\arg\max}~P(x || \mathbf{k^*})
\end{equation}
This TD can be simply found by a greedy search algorithm or a depth first search algorithm.

\section{Evaluation}\label{evaluation}

In this section, we evaluate how well the TRG Model equipped with our training algorithm performs in terms of generating syntactically and semantically correct sentences. We will compare the proposed model with two other recent approaches, focussing on three corpora: Atis \citep{konstas2012uns}, SF-Restaurant \citep{wen2015sem}, and Sumtime-Wind. {\bf Atis} is a corpus containing flight information queries (with 4962 training and 448 testing sample points), taken from the Airline Travel Information Systems Dataset \citep{hemphill1990ati} and transformed into the representation of concept-to-text corpus by \citet{konstas2012uns}. 
{\bf SF-Restaurant} \citep{wen2015sem} is a restaurant information querying corpus, split into training and testing datasets (4075 training and 1024 testing sample points). {\bf Sumtime-Wind} is derived from the Sumtime-Meteo \citep{sripada2002sum} with 500 training and 500 testing sample points. Its Text Descriptions are the wind direction and wind speed 
descriptions, manually extracted from the original wind forecasts. For example, the original forecast
\begin{center}
``W-SW LESS THAN 08 BACKING SSE 06-10 BY AFTERNOON''   
\end{center}
is segmented into two new instances, one of which expresses the concept wind direction while the other expresses wind speed:
\begin{center}
``W-SW LESS THAN 08'' \\
``SSE 06-10''  
\end{center}

\subsection{Experiment Design}\label{expdesign}

We compared our approach with those of \citet{konstas2013glo} and \citet{wen2015sem}. Konstas' approach is a statistical approach driven by a probabilistic context-free grammar (pCFG). Wen's approach is a deep-learning-based approach, namely a Bi-directional LSTM. 

The original plan was to apply all three approaches to all three corpora. However, it proved difficult to apply Wen's method to the Sumtime and ATIS corpora. 
We therefore applied Wen's approach only to his own corpus (i.e. SF-Restaurant), because 
of specific to the dataset preprocessing steps. In Wen's approach, words which express features should be manually
\footnote{\citet{wen2015sem} said the replacement is automatic rather than manual. However, Wen's approach assumes that the concept values must reappear in the TDs. For example, if a feature collection includes $name=zen~yai~thai~restaurant$, then the Text Description has to follow the form: ``... zen yai thai restaurant ...''. Thus, the replacement just finds and replaces words which are the reappearance of the feature value in the texts. For the feature values which do not reappear in the text, Wen's approach manually coded a detailed mapping e.g. ``allow kid -s'' maps to SLOT\_KIDALLOW etc.}
replaced with the concept names from the corpus texts. For example:
\begin{center}
    ``red door caf\'e is a cheap restaurant.''
\end{center}
should be changed to
\begin{center}
    ``SLOT\_NAME is a SLOT\_PRICE restaurant.''
\end{center}
Therefore, Wen's approach is not easy to migrate to other corpora. The TRG model and Konstas' approach were applied on the original SF-Restaurant corpus without the replacement.

We downloaded Wen's code from \footnote{\url{ https://github.com/shawnwun/RNNLG}} and followed the procedures specified there; other than the issue mentioned above, we did not meet any problems. We also downloaded Konstas's code from \footnote{\url{http://www.ikonstas.net/index.php?page=resources}}; since this web site did not contain explicit procedures for training and testing, we asked the authors for clarification, then followed the procedures they kindly provided us with. 

For all three approaches, we first evaluated the syntactic correctness of the sentences generated through grammar checkers. The generated sentences (with the human-written sentences) were judged by the grammar checkers and we counted how many sentences pass without syntax errors. Two general grammar checkers were applied: Language Tool\footnote{\url{https://languagetool.org/}} (LT) -- a commercial grammar checker; and Language Checker\footnote{\url{https://pypi.python.org/pypi/language-check}} (LC) -- an open-source grammar checker. The sentences in Atis and SF-Restaurant are general English so we judged them by LT and LC. However, sentences in Sumtime-Wind are a somewhat formulaeic sub-language that differs from everyday English, so to judge these we built a simple special-purpose grammar checker (denoted as WG), focusing on wind forecasts. 

Secondly, we evaluated syntactic correctness by human judges. Participants were asked to rate the generated sentences on a 10-point scale, rating how well they thought the generated sentences were syntactically correct (with 0 being completely incorrect and 9 being the most accurate). 

Finally, we designed another human experiment to evaluate semantic correctness. We paired each generated sentence with a human-written sentence (as a gold standard). 
We provided the participants with the generated sentences and human-written sentences together, indicating which of the two is the human-written sentence. Then, for each sentence pair, we asked the participants to rate {\it to what extent the generated sentences express the same information as the human-written sentences} on an 11-point rating scale (0 means the two sentences express something entirely different; 10 means they express the exact same information). 

The human experiments (both the one about syntactic correctness and the one about semantic correctness) involved 7 participants, postgraduate students in Computer Science (4 participants) and in Architecture (3 participants). In each of the two experiments, participants were asked to rate as many sentences as possible (using randomly sampled sentences). Each participant in the syntax experiment rated 10-30 sentences from each of the relevant generators in each of the relevant domains (SF-Restaurant, Atis, and Sumtime), and the same is true for the semantics experiment. In total around a thousand examples were rated in each experiment (1364 syntax experiment; 986 semantics experiment).

\begin{figure}[!htbp] 
\centering\includegraphics[width=3.3in]{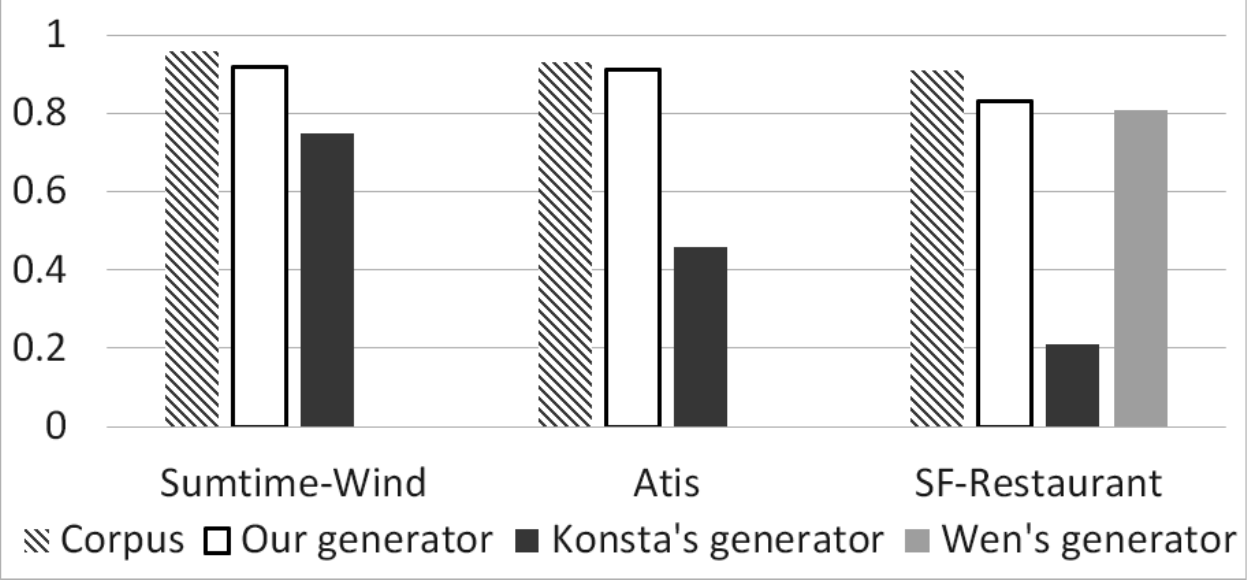} 
\caption{Human evaluation on syntactic correctness.}\label{img1e} 
\end{figure}
\begin{figure}[!htbp] 
\centering\includegraphics[width=3.3in]{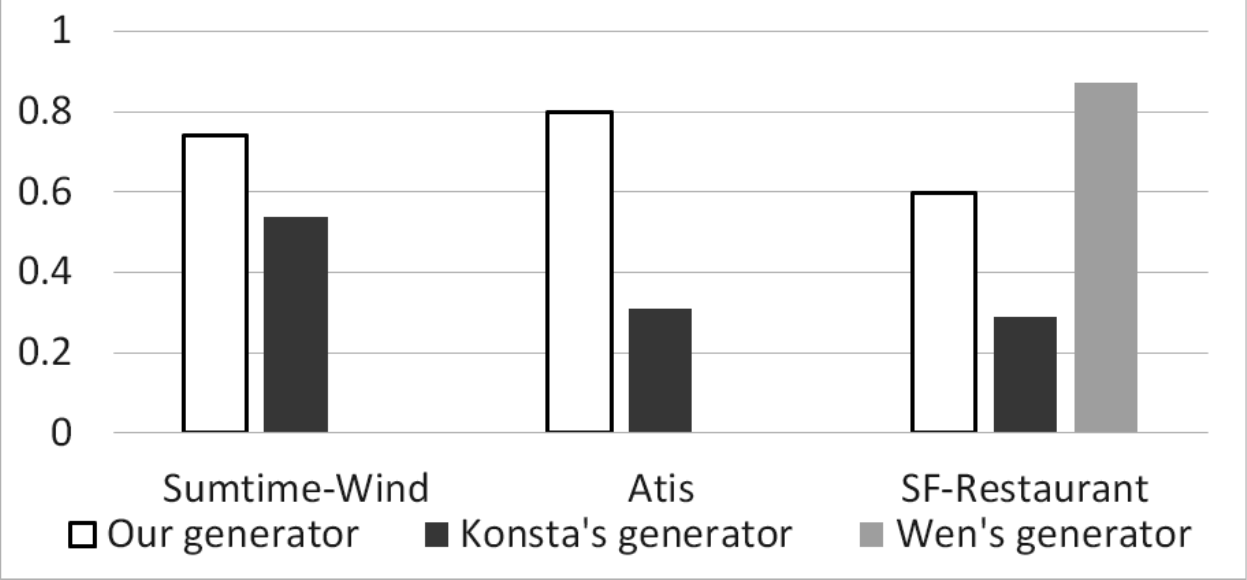} 
\caption{Human evaluation on semantic correctness.}\label{img2e} 
\end{figure}

\begin{table*}[htbp]
\centering
\caption{Syntactic evaluation results (* indicates the performance significantly outcomes the baselines (excluding human-written) in the table with $p<0.01$ by a paired sample $t$-test).}
\label{eval1}
\begin{tabular}{|l|c|c|c|c|c|c|c|c|}
\hline
 & \multicolumn{2}{|c|}{Sumtime-Wind} & \multicolumn{3}{|c|}{Atis} & \multicolumn{3}{|c|}{SF-Restaurant} \\
\hline
 & WG & Human & LT & LC & Human & LT & LC & Human \\
\hline
Human-written & 0.99 &0.96 & 0.97 & 0.95 & 0.93 & 0.95 & 0.96 & 0.90 \\
\hline
Our generator & \textbf{0.98}* & \textbf{0.92}*& \textbf{0.96} & 0.94 & \textbf{0.91}* & \textbf{0.95}* & \textbf{0.96}* & \textbf{0.83} \\
Konstas' generator & 0.54 & 0.75 & 0.92 & \textbf{0.96} & 0.46 & 0.65 & 0.62 & 0.21\\
Wen's generator  & -- & -- & --& --& --& 0.88 & 0.88 & 0.81 \\
\hline
\end{tabular}
\end{table*}

\begin{table*}[htbp]
\centering
\caption{Semantic evaluation results (* indicates the performance significantly outcomes the baselines in the table with $p<0.001$ by a paired sample $t$-test).}
\label{eval3}
\begin{tabular}{|l|c|c|c|}
\hline
 & Sumtime-Wind & Atis & SF-Restaurant \\
\hline
Our generator & \textbf{0.74}* & \textbf{0.80}* & 0.60 \\
Konstas' generator & 0.54 & 0.31 & 0.29 \\
Wen's generator & --&-- & \textbf{0.88} \\
\hline
\end{tabular}
\end{table*}

\subsection{Results}

Syntactic evaluation results are presented in Table \ref{eval1}, and Figure \ref{img1e}, while semantic evaluation results are in Table \ref{eval3}, which is visualized in Figure \ref{img2e}. All the scores from both grammar checkers and human raters were normalized to the decimals between 0 - 1.

The automatic and human results for syntactic evaluation suggest that our generated sentences perform equally well as the human-written sentences. While the overall performance gain of our proposed approach over the approach of Konstas is significant ($p<0.02$ by two-tailed t-test), our approach and Wen's were statistically indistinguishable. 
In the semantic evaluation task, our approach outperformed Konstas' approach ($p<0.03$ by Mann-Whitney U-test for the Likert scales). Wen's model appears to outperform ours, but the results were not statistically significant. It also should be noticed that Wen's approach is trained on the corpus with replacement, which is much easier than the original corpus. 

Both the approaches of us and Konstas performed well on the Sumtime-Wind compared with their performances on SF-Restaurant respectively. It is probably because the sentences in Sumtime-Wind are simpler than in SF-Restaurant with just two classes of concepts.

In the experiments, Konstas' approach gains surprisingly low scores. We believe this might be for the following reason. In Konstas' experiments, both a gold standard (the human-written sentences) and a basic baseline (sentences generated by a unigram model) are involved, and they were provided to the human participants together with the sentences generated by Konstas' approach. Because sentences of the unigram baseline are almost unreadable; they define a very low baseline, to distinguish with it, participants would produce the higher scores to Konstas' approach. However, in our experiment, we did not involve a basic baseline, but we did involve the gold standard (the human-written sentences). Consequently, our participants tended to rate them by the stricter standard.

%
\section{Discussion}\label{discussion}

In this paper, we have presented a fully statistical approach to NLG which is based on a extraction of placeholders and schemas. Evaluation results on three different corpora suggest that our approach performs well, both in terms of syntactic correctness (based on both automatic metrics and human judgements) and in terms of semantic correctness (based on human judgments). The method generally outperformed that of \citet{konstas2013glo}, even on the ATIS corpus, for which their method was originally developed. Comparing the performance of our method against \citet{wen2015sem} is proved to be more difficult, because Wen's method could only be applied to one of the three corpora that we studied; on this one corpus (i.e., the one developed by the authors themselves), however, Wen's method outperformed ours. 

Researchers have employed statistically inspired approaches to NLG for a number of years now, but we have argued that many of the proposed approaches have not been {\em fully} statistical. The approach of \cite{belz2008aut}, for example, relied on a pCFG that was trained from a hand-written CFG. In some cases, as we have seen in the section of experiment design, even the approach of \cite{wen2015sem}, which aims to be purely statistical, appears to rely on alignment procedures that are not fully specified. The approach proposed in the present paper does away with all handcoding and offers a fully statistical approach to NLG.

We believe, moreover, that it is an important advantage of our approach that it is {\em inspectable} to a much greater extent than many of its competitors. In particular, it is possible to inspect what words (fragments) correspond to what placeholders, and to change them to modify the behaviour of the system if this is desired; similarly, it is possible to scrutinise the set of schemas, and to modify one or more schemas in light of new evidence.
After training, what TRG model learns is the generation rules -- the schemas and fragments, which could still be easily modified by human users. Through reading the schemas and fragments, human users can generally understand what sentences the rules result. If particular sentences should be generated or be avoid to generate, manually adding or removing the relative schemas and fragments is enough to do so. 
A lack of inspectability is often cited as one of the main drawbacks of approaches based on Neural Networks \cite{techtalks.tv}. After training, what the neural-network-based approaches learn is massive weights of matrices. Human users cannot trace what particular sentences can be generated according to the matrices. Consequently, to modify a trained approach is impossible.

A limitation of the work presented here is that the training algorithm we proposed does not use the training data efficiently.
The TRG algorithm is very harsh on the assumption of placeholder equality. Only the same placeholders of the same schema can share a small corpus. This strict criterion is to ensure the syntactic correctness of the generated TD, because words expressing the same feature may have different syntactic roles (i.e. word forms, e.g. none, adj, adv, etc.). However, this strict criterion also makes the division of small corpora being too fine. When the training corpus is small, the placeholders in the unpopular schema cannot obtain the acceptable size of the small corpus.


Although the possibility of lacking data potentially limits the diversity of TD generated by our algorithm under small dataset, the three corpora used in our evaluation are large enough, so our algorithm still performs. The TRG model can still learn enough schemas and fragments to perform the task of the TD generation. Meanwhile, when the selectors selects schemas and fragments, it gives priority to the schemas and fragments with more occurrences. So the limitation does not (or rarely) disturb our evaluation.


\section{Conclusion}

This paper proposed a purely statistical NLG approach -- TRG model. Given a concept-to-text corpus, the approach can automatically analyse and learn how TDs in the corpus express the corresponding data (feature collections), and generates new TDs by imitating the corpus TDs. 
The algorithm generates TDs based on the split-reassemble strategy. We first decompose the TDs in the corpus into fragments and schemas by calculating the fragment-feature alignment. In the aligning process, we calculate what fragments express what features, and align the most appropriate fragments to the corresponding features. When the model generates a TD for a new feature collection, the model selects the schema and fills the placeholders of the schema with fragments to get the complete TD. In the evaluation, we verified that the generated TD has good ideographic ability and syntactic correctness based on three different data-text corpora. In the future, we will continue to improve the model and the efficiency of the schema extraction algorithm, such as trying to detect and merge small corpora so that each placeholder gets more sufficient candidate fragments.

This approach combines the advantages of machine-learning and rule-based NLG approaches. It is considered as the combination of a statistical method for rule extraction with a rule-based generation process, and the learnt schemas and the semantically meaningful fragments can be considered as the generation rules.
This model presents the automatic training process and also keeps the transparency. The advantages of automation allow our program to learn the given corpus without any process of manually analysing and summarising the rules or templates for generation like what rule-based NLG programs have to do. 
Transparency makes the learnt results of the approach (i.e. schemas and the semantically meaningful fragments) easily modified by hand, so the performance of TRG model can be easily predicted and corrected. This ability is especially important for NLG programs to become commercial software, and this is exactly what the machine-learning NLG approaches do not keep.

\bibliographystyle{authordate1}
\newcommand{\newblock}{}
\bibliography{main}

\label{lastpage}

\end{document}